\documentclass[10pt,twocolumn,letterpaper]{article}

\usepackage{cvpr}
\usepackage{times}
\usepackage{epsfig}
\usepackage{graphicx} 
\graphicspath{{figure/}}
\usepackage{enumitem}
\usepackage{caption}
\usepackage{dsfont}
\usepackage{wrapfig}
\usepackage{amsfonts,amsmath,amssymb,amsthm}
\usepackage{bm,nicefrac}

\usepackage[pagebackref=true,breaklinks=true,letterpaper=true,colorlinks,bookmarks=false]{hyperref}

\cvprfinalcopy 

\newcommand{\setmode}[1]{\def\mode{#1}}
\setmode{draft} 
\long\def\IGNORE#1{} \long\def\COMMENT#1{}

\ifthenelse{\equal{\mode}{draft}} { 
    \def\authornote#1#2#3{{\textcolor{#2}{\textsl{\small#1:[*#3*]}}}}
    } {}

\ifthenelse{\equal{\mode}{final}} {
    \def\authornote#1#2#3{}
    \typeout{************* MODE: Final}
    } {}

\cvprfinalcopy

\begin{document}

\title{\vspace{-9mm}
Novel View Synthesis of Dynamic Scenes with Globally Coherent Depths \\ from a Monocular Camera }

\author{
Jae Shin Yoon$^\dagger$
\hspace{5mm}Kihwan Kim$^\sharp$
\hspace{5mm}Orazio Gallo$^\sharp$
\hspace{5mm}Hyun Soo Park$^\dagger$
\hspace{5mm}Jan Kautz$^\sharp$
\\
\hspace{-20mm}$^\dagger$University of Minnesota
\hspace{38mm}
$^\sharp$NVIDIA \\
}
\pagenumbering{gobble}
\thispagestyle{empty}

\twocolumn[{%
\maketitle
	\begin{center}
		\centering
		\vspace{-6mm}
	 \includegraphics[width=.94\linewidth]{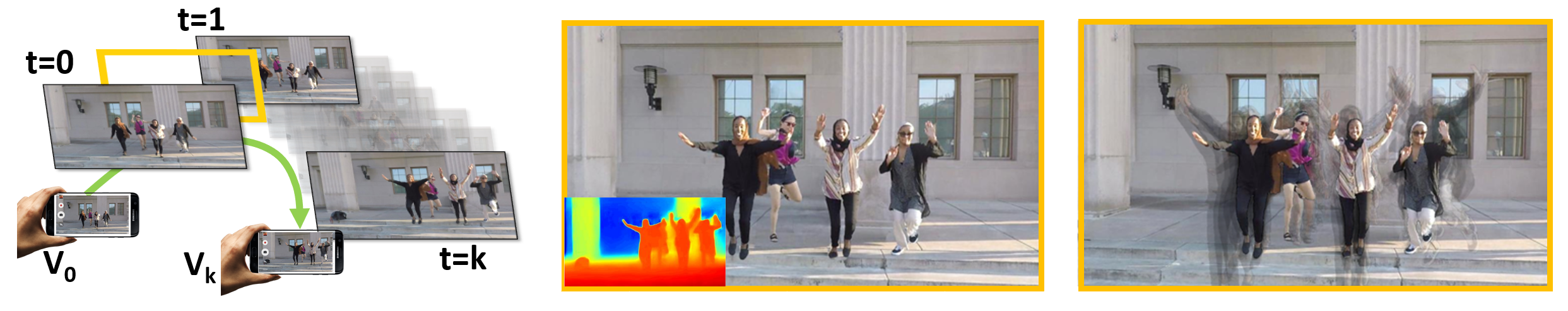}
	 \vspace{-2mm}
	\captionof{figure}{\textbf{Dynamic Scene View Synthesis:} (Left) A dynamic scene is captured from a monocular camera from the locations $\mathbf{V}_0$ to $\mathbf{V}_k$. Each image captures people jumping at each time step ($\mathbf{t}=0$ to $\mathbf{t}=k$). (Middle) A novel view from an arbitrary location between $\mathbf{V}_0$ and $\mathbf{V}_1$ (denoted as an orange frame) is synthesized with the dynamic contents observed at the time  $\mathbf{t} =k$. The estimated depth at $\mathbf{V}_k$ is shown in the inset. (Right) For the novel view (orange frame), we can also synthesize the dynamic content that appeared across any views in different time (traces of the foreground in each time step are shown). More results are shown in Sec.~\ref{sec:exp} and the supplementary document and video.
	}
	\label{fig:teaser}

\end{center}	
\label{fig:teaser_main}
}]

\begin{abstract}
\vspace{-9mm}
This paper presents a new method to synthesize an image from arbitrary views and times given a collection of images of a dynamic scene.
A key challenge for the novel view synthesis arises from dynamic scene reconstruction where epipolar geometry does not apply to the local motion of dynamic contents. To address this challenge, we propose to combine the depth from single view (DSV) and the depth from multi-view stereo (DMV), where DSV is complete, i.e., a depth is assigned to every pixel, yet view-variant in its scale, while DMV is view-invariant yet incomplete. Our insight is that although its scale and quality are inconsistent with other views, the depth estimation from a single view can be used to reason about the globally coherent geometry of dynamic contents. We cast this problem as learning to correct the scale of DSV, and to refine each depth with locally consistent motions between views to form a coherent depth estimation. We integrate these tasks into a depth fusion network in a self-supervised fashion. Given the fused depth maps, we synthesize a photorealistic virtual view in a specific location and time with our deep blending network that completes the scene and renders the virtual view. We evaluate our method of depth estimation and view synthesis on diverse real-world dynamic scenes and show the outstanding performance over existing methods.
\vspace{9mm}


\end{abstract}

\section{Introduction}
\label{sec:intro}

Novel view synthesis~\cite{Chen93} is one of the core tasks in computer vision and graphics, and has been used for many visual effects and content creation applications such as cinemagraph~\cite{Bai:2013:ACP:2600890.2600894,Liao:2013:AVL:2461912.2461950}, video stabilization~\cite{liu2013bundled,Kopf:2014:FHV:2601097.2601195}, and bullet time visual effect~\cite{zitnick2004high}. In this paper, we focus on view synthesis of dynamic scenes observed from a moving monocular camera as shown in Figure~\ref{fig:teaser}.
Until now, most of existing view synthesis methods are largely limited to static scenes~\cite{Chen93,zitnick2004high,choi2018extreme,Flynn_2019_CVPR,zhang2009consistent,zhou2018stereo} because they commonly rely on geometric assumptions:
in principle, dynamic visual content such as people, pets, and vehicles are considered \textit{outliers} despite being a major focus in videography on social media and otherwise. 

Our problem shares the challenge of dynamic scene reconstruction: recovering the underlying 3D geometry of dynamic contents from a moving monocular camera is fundamentally ill-posed~\cite{park20113d}. 
We address this challenge by leveraging the following complementary visual and motion cues. 
(1) Multi-view images can be combined to reconstruct incomplete yet view-invariant static scene geometry\footnote{Its fixed scale chosen from SfM pipeline is consistent across different views from initial triangulation~\cite{hartley:2004}.}, which enables synthesizing a novel view image of static contents in a geometrically consistent way. 


\noindent(2) Relative depth predicted from a single image provides view-variant~\cite{bian2019unsupervised} yet complete dynamic scene geometry, which allows enforcing locally consistent 3D scene flow for the foreground dynamic contents.



\IGNORE{

\begin{figure}[t]
	\begin{center}
\vspace{-5mm}
\includegraphics[width=1\linewidth]{./figure/scale_inconsistency3.pdf}
	\end{center}
\vspace{-5mm}
    \caption{\small Image warping from a view to the other view using the depth from single view (DSV). The purple color shows the image misalignment of the warped image with ground-truth originated from the depth scale inconsistency across the views.}
    \label{fig:incosistency}
    \vspace{-.3cm}
\end{figure}

}

We combine these cues by learning a nonlinear scale correction function that can upgrade a time series of single view geometries to form a coherent 4D reconstruction. 
To disambiguate the geometry of the foreground dynamic contents, we find their simplest motion description in 3D (i.e., slow and smooth motion~\cite{valmadre:2012,Russell:2011:EBM:2191740.2191962}), which generates minimal stereoscopic disparity when seen by a novel view~\cite{Ballan:2010:UVR:1778765.1778824}.



\begin{figure}
	\begin{center}
    \includegraphics[width=0.99\linewidth]{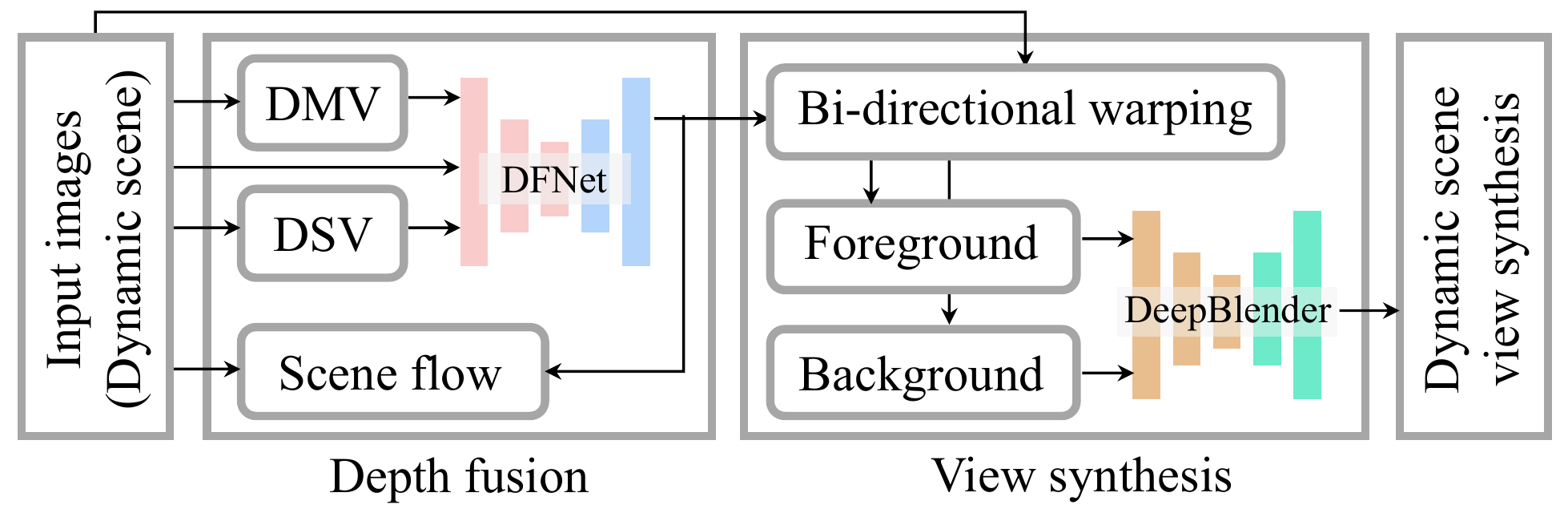}
	\end{center}	
	\vspace{-.6cm}
	\caption{\small Images of a dynamic scene are used to predict and estimate the depth from single view (DSV) and the depth from multi-view stereo (DMV). Our depth fusion network (DFNet) fuses the individual strengths of DSV and DMV (Sec.~\ref{depth_fusion}) to produce a complete and view-invariant depth by enforcing geometric consistency. The computed depth is used to synthesize a novel view and our DeepBlnder network refines the synthesized image (Sec.~\ref{deep_blender}). 
    }
	\label{fig:overview}
	\vspace{-.3cm}
\end{figure}

We model the scale correction function using a depth fusion network that takes input images, view-variant depth from single view (DSV), and incomplete yet view-invariant depth from a multi-view stereo (DMV) algorithm, and outputs complete and view-invariant depth. The network is self-supervised by three visual signals: (i) the static regions of the DSV must be aligned with a DMV; (ii) the output depth of dynamic regions must be consistent with the relative depth of each DSV; and (iii) the estimated scene flow must be minimal and locally consistent. With the predicted depths that are geometrically consistent across views, we synthesize a novel view using a self-supervised rendering network that produces a photorealistic image in the presence of missing data with adversarial training. An overview of our pipeline is shown in Figure~\ref{fig:overview}.

We show that the novel view synthesis with our depth prediction method is highly effective in generating an unseen image. Further, the rendering network seamlessly blends foreground and background, which outperforms existing synthesis approaches quantitatively and qualitatively.



Our key contributions are as follows: 
\begin{itemize}[topsep=1pt,itemsep=1pt,partopsep=0pt, parsep=0pt]
\item A novel depth fusion network that models a scale correction function, which completes the depth maps of view-invariant dynamic scene geometry with locally consistent motions.
\item A rendering network that combines foreground and background regions in a photorealistic way using adversarial training.
\item A real-world dataset captured by fixed baseline multi-view videos and corresponding benchmark examples for dynamic scene view synthesis.
\end{itemize}

\section{Related Work}
For the view synthesis of a dynamic scene from images with baselines, the depth and foreground motion from each view need to be consistent across the views. Here we review view synthesis, depth estimation, and scene reconstruction techniques, and discuss the relations to our method.

\begin{figure*}[ht]
	\begin{center}
		\includegraphics[width=.94\linewidth,trim=2pt 1mm 3mm 2pt,clip]{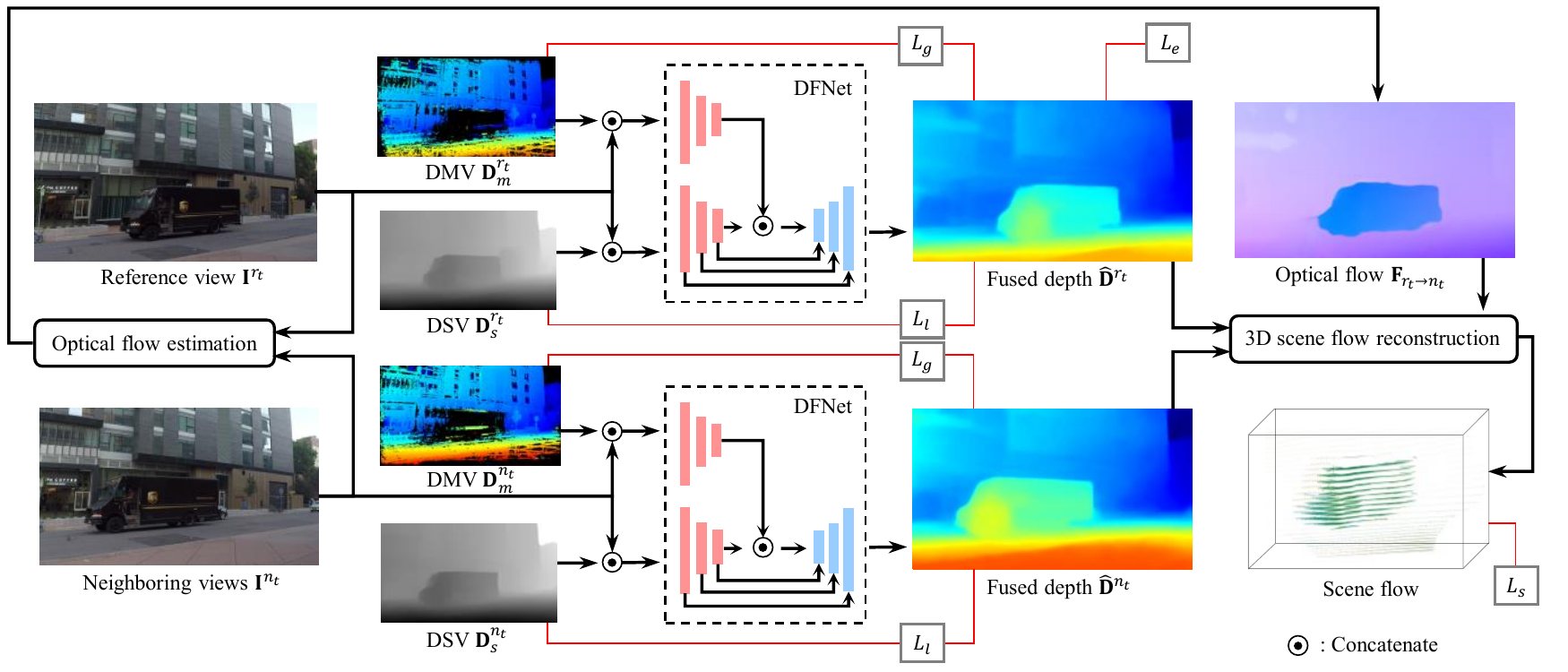}
	\end{center}	
	\vspace{-4mm}
	\caption{\small Depth Fusion Network (DFNet) predicts a complete and view-invariant depth map by fusing DSV and DMV with the image. DFNet is self-supervised by minimizing the background depth consistency with DMV ($L_g$), the relative depth consistency with DSV ($L_l$), 3D scene flow ($L_s$), and spatial irregularity ($L_e$).}
	\label{fig:gldfnet_all}
\end{figure*}

\noindent\textbf{Novel View Synthesis} The problem of novel view synthesis is strongly tied to multiview 3D reconstruction as it requires transporting pixels across views through the geometry of scenes. For a static scene, multiview geometry~\cite{hartley:2004} applies, which allows triangulating correspondences given the calibrated cameras. This leads to dense reconstruction (multiview stereo~\cite{furukawa:2010}), which allows continuous view synthesis using a finite number of images~\cite{frahm2010building}. In principle, the triangulation requires a correspondence from at least two views. This requirement often cannot be met in particular for occluded regions, which results in an incomplete view synthesis. Such issue escalates as the baseline between images increases. Various scene priors have been used to mitigate this issue. A scene that can be expressed as a set of planar surfaces can be reconstructed even with a single view image~\cite{sinha:2009,liu2019planercnn}, and a scene with known object categories can be reconstructed with the shape priors~\cite{hoiem:2005}. For dynamic scene view synthesis, synchronized multiple cameras are used where the same geometric principle can be applied~\cite{zitnick2004high,lombardi2018deep,jiang20123d}. Recently, single view depth prediction that is learned from a large image repository is used to complete the scene geometry~\cite{Philip:2018:PMI:3190834.3190846,li2018megadepth,lasinger2019towards},
and even enables to extrapolate the views beyond the range of camera motion~\cite{srinivasan19,choi2018extreme}. More recent approach, which predicts the depth from single view with human specific priors, realizes the view synthesis of dynamic scene of moving people from a monocular camera~\cite{li2019learning}. Our approach is inspired by learning based scene completion~\cite{flynn2016deepstereo,Philip:2018:PMI:3190834.3190846,zhou2018stereo,li2019learning} while applying to a dynamic scene with geometric consistency without any category-specific priors.
















\noindent\textbf{Monocular Dynamic Scene Reconstruction} Dynamic scene reconstruction using a moving monocular camera, without a prior assumption of scenes, is very challenging and ill-posed, similar to reconstructing a scene with a single view image. For a temporal prior, trajectory triangulation extends the concept of point triangulation by representing scene trajectories using a family of algebraic groups, e.g., line/conic~\cite{avidan:2000}, homography tensors on the plane~\cite{shashua:2000,wexler:2000}, polynomials~\cite{kaminski:2004}, and discrete cosine transform basis~\cite{akhter:2008,park20113d}. For spatial prior, a shape can be expressed by a linear combination of compact shape basis vectors~\cite{bregler:2000}, which is highly effective to describe constrained deformation such as face. A key challenge is learning shape basis vectors for unknown objects, which requires additional spatial priors such as orthogonality of basis vectors~\cite{xiao:2004}, temporal smoothness~\cite{torresani:2002,olsen:2007,valmadre:2012}, articulation constraint using joint subspace~\cite{yan:2005}, local piecewise rigidity~\cite{fayad:2010}, and learning from training data~\cite{1902.10840}. For completeness, image regions are reconstructed independently using shape basis~\cite{Russell:2014} or local patches~\cite{kumar:2017}, which can be stitched together to form complete scene reconstruction. Further, the spatial and temporal priors can be combined to produce dense correspondences, resulting in complete 4D reconstruction~\cite{ji:2016}. Humans are a special case of spatial constraints, which allow markerless motion capture from a monocular camera~\cite{xiang:2019,SMPL-X:2019,Alldieck:2019}. Unlike the explicit spatial priors, our work makes use of general geometric priors and motion constraint to reconstruct a complete and view-invariant geometry of general dynamic scenes, which allows us to generate realistic spatio-temporal view synthesis.

\section{Approach}\label{approach}


We cast the novel view synthesis problem as image warping from input source views to a virtual view using underlying 4D reconstruction, i.e., 
\begin{align}
\mathbf{J}^v(W_{r\rightarrow v}(\mathbf{x})) = \mathbf{I}^r(\mathbf{x}),    
\end{align}
where $\mathbf{J}^v$ is the synthesized image from an arbitrary virtual view $v$ ($v$ can be a source viewpoint),
$W_{r \rightarrow v}$ is a warping function, and $\mathbf{I}^r$ is the $r^\text{th}$ source image. 

For view synthesis of static scene, the warping function can be described as:
\begin{align}
    \mathbf{y} = W_{r\rightarrow v}(\mathbf{x};\mathbf{D}^r,\Pi^r, \Pi^v),\label{Eq:static}
\end{align}
where $\Pi^r$ and $\Pi^v$ are the projection matrices at the $r^{\rm th}$ and $v^\text{th}$ viewpoints. The warping function forms the warped coordinates $\mathbf{y}$ by reconstructing the view-invariant 3D geometry using the depth ($\mathbf{D}^r$) and projection matrix at the $r^\text{th}$ viewpoint, and projecting onto the $v^\text{th}$ viewpoint.
For instance, this warping function can generate the $i^\text{th}$ source image from the $j^\text{th}$ source image, i.e., $\mathbf{I}^{i}(W_{j\rightarrow i}) = \mathbf{I}^{j}$.

For view synthesis of dynamic scene, the warping function can be generalized to include the time-varying geometry using the depth $\mathbf{D}^{r_t}$, i.e.,
\begin{align}
    \mathbf{y} = W_{r_t\rightarrow v}(\mathbf{x};\mathbf{D}^{r_t},\Pi^r, \Pi^v),
\end{align}
where $r_t$ is the time dependent view index, and $t$ is the time instant. Note that for a moving monocular camera, the view is a function of time. Unlike the static scene warping $W_{r\rightarrow v}$ in Eq.~(\ref{Eq:static}), we cannot synthesize $i^{\rm th}$ source image from the $j^{\rm th}$ source image because of the time-varying geometry $\mathbf{D}^{r_t}$, i.e., $\mathbf{I}^{i}(W_{j\rightarrow i}) \neq \mathbf{I}^{j}$.

With these two warping functions, the dynamic scene view synthesis can be expressed as:
\begin{align}
    \mathbf{J} = \phi\left(\left\{\mathbf{J}^v\left(W_{r\rightarrow v}\right)\right\}_r, \mathbf{J}^{v,t}\left(W_{r_t\rightarrow v}\right); \mathcal{M}^{v}\right),\label{Eq:syn}
\end{align}
where $\{\mathbf{J}^v(W_{r\rightarrow v})\}_r$ is a set of static scene warping from all source viewpoints, and $\mathbf{J}^{v,t}(W_{r_t\rightarrow v,t})$ is the warping of dynamic contents from the source image of the $t^{\rm th}$ time instant. $\mathcal{M}^{v}$ is the set of the coordinates belonging to dynamic contents. $\phi$ is the rendering function that refines the warped images to complete the view synthesis.


In Eq.~(\ref{Eq:syn}), two quantities are unknowns: the depth from each source view $\mathbf{D}^{r_t}$ and the rendering function $\phi$. We formulate these two quantities in Sec.~\ref{depth_fusion} and Sec.~\ref{deep_blender}.

\subsection{Globally Coherent Depth from Dynamic Scenes}\label{depth_fusion}
Our conjecture is that there exists a scale correction function that can upgrade a complete view-variant depth $\mathbf{D}_s^{r_t}$ from the single view prediction (DSV) to the depth of the view-invariant 3D geometry $\widehat{\mathbf{D}}^{r_t}$:
\begin{align}
    \widehat{\mathbf{D}}^{r_t} =\psi(\mathbf{D}^{r_t}_s), 
\end{align}
where $\psi$ is the scale correction function. 
Ideally, when a scene is stationary, the upgraded depth is expected to be identical to the depth $\mathbf{D}_m^r$ from view-invariant geometry, e.g., depth from multiview stereo (DMV), with uniform scaling, i.e., $\mathbf{D}_m^r = \psi(\mathbf{D}_s)= \alpha \mathbf{D}_s + \beta$ where $\alpha$ and $\beta$ are scalar and bias. When a scene is dynamic, the linear regression of such scale and bias is not applicable. We learn a nonlinear scale correction function that possesses the following three properties.




First, for the static scene, the upgraded depth approximates DMV:
\begin{align}
    \mathbf{D}^r_m(\mathbf{x}) \approx \psi\left(\mathbf{D}_s^{r_t}(\mathbf{x})\right)~~~{\rm for}~ \mathbf{x} \notin \mathcal{M}^{r_{t}}, 
\label{property:loss1}
\end{align}
where $\mathbf{x}$ is the coordinate of pixels belonging to the static background.

Second, for the dynamic contents, the upgraded depth preserves the relative depth from DSV:
\begin{align}
    g\left(\mathbf{D}^{r_t}_s(\mathbf{x})\right) \approx g\left(\psi\left(\mathbf{D}^{r_t}_s(\mathbf{x})\right)\right)~~~{\rm for}~ \mathbf{x} \in \mathcal{M}^{r_{t}},
\label{property:loss2}
\end{align}
where $g$ measures the scale invariant relative gradient of depth, i.e.,
\begin{align}
g(\mathbf{D};\mathbf{x},\Delta\mathbf{x})={\frac{\mathbf{D}(\mathbf{x}+\Delta\mathbf{x})-\mathbf{D}(\mathbf{x})}{|\mathbf{D}(\mathbf{x}+\Delta\mathbf{x})|+|\mathbf{D}(\mathbf{x})|}}.
\label{scale_invariant_loss}
\end{align}
We use multi-scale neighbors $\mathbf{x}+\Delta\mathbf{x}$ to constrain local and global relative gradients.


Third, 3D scene motion induced by the upgraded depths is smooth and slow~\cite{valmadre2012general}, i.e., minimal scene flow: 
\begin{align}
\label{property:loss3}
\mathbf{p}(\mathbf{x}; \mathbf{D}^{r_t}, \Pi^{r_t}) \approx \mathbf{p}(F_{r_t\rightarrow n_t}(\mathbf{x}); \mathbf{D}^{n_t}, \Pi^{n_t}),
\end{align}
where $F_{r_t\rightarrow n_t}$ is the optical flow from the $r_t^{\rm th}$ to $n_t^{\rm th}$ source images. $\mathbf{p}(\mathbf{x}; \mathbf{D}) \in \mathds{R}^3$ is the reconstructed point in the world coordinate using the depth $\mathbf{D}$:
\begin{align}
    \mathbf{p}(\mathbf{x}; \mathbf{D}, \Pi) =  \psi\left(\mathbf{D}(\mathbf{x})\right)\mathbf{R}^\mathsf{T}\mathbf{K}^{-1}\widetilde{\mathbf{x}} + \mathbf{C}
\end{align}
where $\widetilde{\mathbf{x}}$ is the homogeneous representation of $\mathbf{x}$, and $\mathbf{R}\in SO(3)$, $\mathbf{C}\in \mathds{R}^3$, and $\mathbf{K}$ are the camera rotation matrix, camera optical center, and camera intrinsic parameters from the projection matrix $\Pi$.

\noindent\textbf{Depth Fusion Network (DFNet)} We enable the scale correction function $\psi$ using a depth fusion network that takes as input DSV, DMV, and image $\mathbf{I}^{r_t}$:
\begin{align}
    \widehat{\mathbf{D}}^{r_t} = \psi\left(\mathbf{D}_s^{r_t}, \mathbf{D}_m^{r_t}, \mathbf{I}^{r_t}; \mathbf{w}\right),
\end{align}
where the network is parametrized by its weights $\mathbf{w}$. To learn $\mathbf{w}$, we minimize the following loss:
\begin{align}
    L(\mathbf{w}) = L_g + \lambda_l L_l + \lambda_s L_s+\lambda_e L_e,
\end{align}
where $\lambda$ controls the importance of each loss. $L_g$ measures the difference between DMV and the estimated depth in Eq.~(\ref{property:loss1}) for static scene: 
\begin{align}
L_g= \|\widehat{\mathbf{D}}^{r_t}(\mathbf{x})-\mathbf{D}^{r_t}_m(\mathbf{x})\|~~~{\rm for}~~ \mathbf{x} \notin \mathcal{M}^{r_{t}},
\nonumber
\end{align}
\noindent$L_l$ compares the scale invariant depth gradient between DSV and the estimated depth in Eq.~(\ref{property:loss2}):
\begin{align}
L_l= \|g(\widehat{\mathbf{D}}^{r_t}(\mathbf{x}))-g(\mathbf{D}^{r_t}_s)(\mathbf{x})\|~~~{\rm for}~~~\mathbf{x} \in \mathcal{M}^{r_{t}},
\nonumber
\end{align}
and $L_s$ minimize the induced 3D scene motion for entire pixel coordinates in Eq.~(\ref{property:loss3}):
\begin{align}
L_s= \|\mathbf{p}(\mathbf{x}; \mathbf{D}^{r_t}, \Pi^{r_t}) - \mathbf{p}(F_{r_t\rightarrow n_t}(\mathbf{x}); \mathbf{D}^{n_t}, \Pi^{n_t})\|. 
\nonumber
\end{align}
\noindent In conjunction with self-supervision, we further minimize the Laplacian of the estimated depth as regularization, i.e., 
\begin{align}
L_e= \|\nabla^2\widehat{\mathbf{D}}^{r_t}(\mathbf{x})\|^2+\lambda_f\|\nabla^2\widehat{\mathbf{D}}^{r_t}(\bar{\mathbf{x}})\|^2
\end{align}
where $\mathbf{x}\notin\mathcal{M}^{r_t}$, $\bar{\mathbf{x}}\in\mathcal{M}^{r_{t}}$, and $\lambda_f$ balances the spatial smoothness between the static and dynamic regions.

The overview of our self-supervision pipeline and the network architecture are described in Figure~\ref{fig:gldfnet_all}. DFNet extracts the visual features from DSV and DMV using the same encoder in conjunction with the image. With the visual features, DFNet generates a complte and view invariant depth map that is geometrically consistent. To preserve the local visual features, skip connections between the feature extractor and depth generator are used.

\begin{figure}[t]
	\begin{center}
		\includegraphics[width=0.47\textwidth]{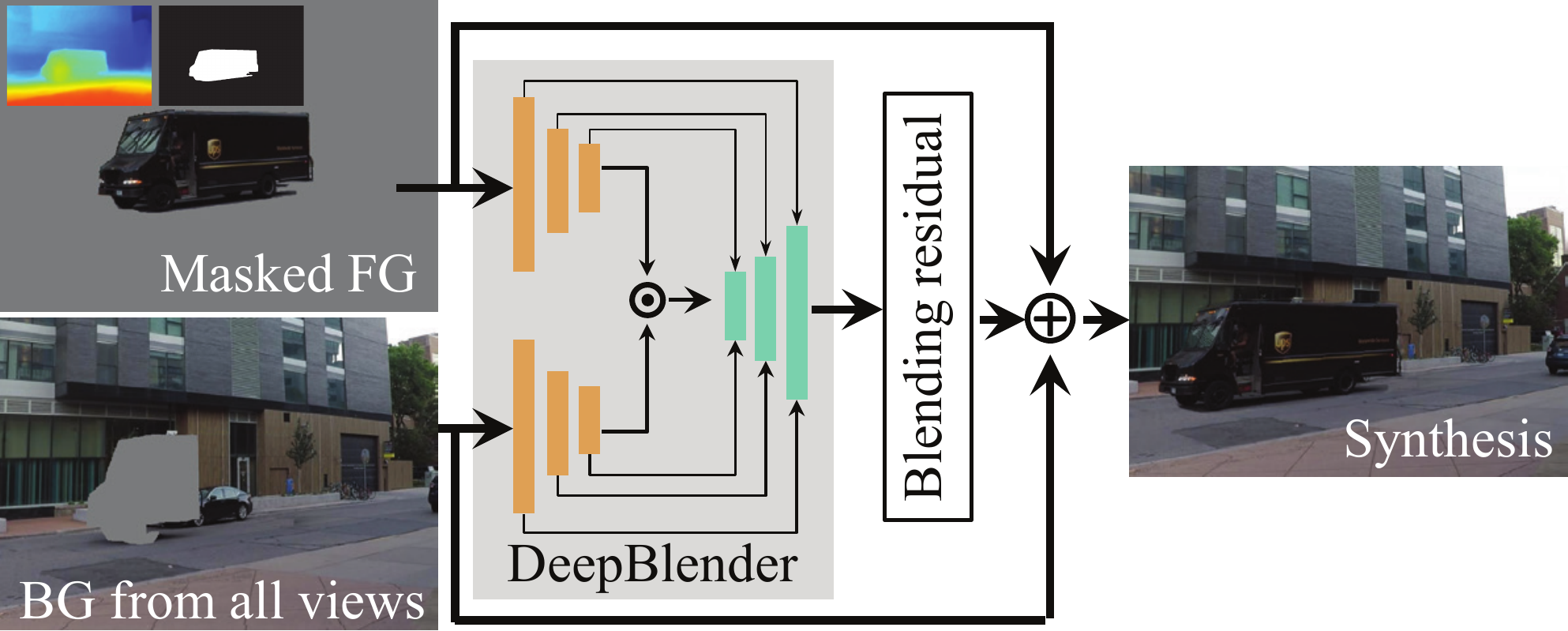}
	\end{center}	
	\vspace{-5mm}
	\caption{\small View synthesis pipeline: Given the warped foreground (FG) and background (BG) through the depths and masks, we complete the dynamic scene view synthesis using a rendering network called DeepBlender that predicts the missing region and refines the artifacts. 
	}
	\label{fig:deepblender}
	\vspace{-.2cm}
\end{figure}


\subsection{Dynamic Scene View Synthesis}\label{deep_blender}
Given a set of warped static scenes from all source views $\{\mathbf{J}^v\}_r$, we construct a global background $\mathbf{J}^v_{*}$ based on the baseline between the virtual and source cameras, i.e., assign the pixel value from the warped source view that has the shortest baseline with virtual camera. With $\mathbf{J}^v_{*}$ and the warped dynamic contents $\mathbf{J}^{v,t}$ from a single time instant, we model the synthesis function $\phi$ in Eq.~(\ref{Eq:syn}) as follows: 
\begin{align}
\phi(\mathbf{J}^{v}_{*}, \mathbf{J}^{v,t};\mathcal{M}^{v})=\mathbf{J}^{v}_{*}(\mathbf{x})+\mathbf{J}^{v,t}(\mathbf{y})+\widetilde{\phi}_{\theta}(\mathbf{J}^{v}_{*},\mathbf{J}^{v,t}),
\end{align}
where $\mathbf{x}\notin\mathcal{M}^{v,t}$ and $\mathbf{y}\in\mathcal{M}^{v,t}$.
$\widetilde{\phi}_{\theta}$ is the blending residual that fills the missing regions (unlike a static scene, there exists the regions that are not seen by any source views for a dynamic scene) and refines the synthesized image. We model this blending residual $\widetilde{\phi}_{\theta}$ using our rendering network.







\noindent\textbf{DeepBlender Network} The DeepBlender predicts the blending residual $\widetilde{\phi}_{\theta}$ from the inputs of a warped dynamic scene $\mathbf{J}^{v,t}$ and a globally modeled static scene $\mathbf{J}^{v}_{*}$ as shown in Figure~\ref{fig:deepblender}. It combines visual features extracted from $\mathbf{J}^{v,t}$ and $\mathbf{J}^{v}_{*}$ to form a decoder with skip connections. We learn this rendering function using source images with self-supervision. Each image is segmented into background and foreground with the corresponding foreground mask. We synthetically generate the missing regions near the foreground boundary and image border, and random pixel noises across the scenes. From the foreground and background images with missing regions and pixel noises, the DeepBlender is trained to generate the in-painting residuals. We incorporate an adversarial loss to produce photorealistic image synthesis:
\begin{align}
L(\mathbf{w}_{\theta})= L_{\rm rec}+\lambda_{\rm adv}L_{\rm adv},
\label{gloss}
\end{align}
where $L_{\rm rec}$ is the reconstruction loss (difference between the estimated blending residual and ground truth), and $L_{\rm adv}$ is the adversarial loss~\cite{pathakCVPR16context}. The overview of our view synthesis pipeline is described in Figure~\ref{fig:deepblender}.

\section{Implementation Details}
DFNet is pre-trained on a synthetic dataset~\cite{lv2018learning} (which provides ground-truth optical flow, depth, and foreground mask) for better weight initialization during the self-supervision. To simulate the characteristic of the real data from synthetic, we partially remove the depth around the foreground region and add the depth noise across the scenes with 5\% tolerance of the variance at every training iteration. The same self-supervision loss as Eq.~\ref{gloss} is used to pre-train the network. To avoid the network depth scale confusion, we use the normalized inverse depth~\cite{lasinger2019towards} for both DMV and DSV and recover the scale of the fused depth based on the original scale of DMV. To obtain DSV and DMV, we use existing single view prediction~\cite{lasinger2019towards} and multiview stereo method~\cite{schonberger2016pixelwise}. In Eq.~(\ref{scale_invariant_loss}), we use five multi-scale neighbors, i.e., $\Delta\textbf{x}=\{1,2,4,8,16\}$ to consider both local and global regions. We use PWCNet~\cite{Sun2018PWC-Net} to compute the optical flow in Eq.~(\ref{property:loss3}), where the outliers were handled by forward-backward flow consistency. When enforcing the scene flow loss, we use $\pm2$ neighboring camera views, i.e., $n_t=r_t\pm2$. We extract the foreground mask using interactive segmentation tools~\cite{rother2004grabcut}. The foreground masks are manually specified for all baselines in the evaluation, while existing foreground segmentation approaches~\cite{Qin_2019_CVPR} can be used as a complementary tool as shown in Figure~\ref{fig:mask}.

We also pre-train the DeepBlender using video object segmentation dataset~\cite{Perazzi2016}. To create the synthetic residual, we randomly generate the seams and holes around the foreground using mask morphology and superpixel, and remove one side of the image boundary up to 30-pixel thickness. The loss in Eq.~\ref{gloss} is used for pre-training as well. When we warp an image to a virtual view, we check bidirectional warping consistency to prevent the pixel holes. For each image warping, we refine the depth using the bilateral weighted median filters~\cite{zhang2014100+}. As shown in Figure~\ref{fig:deepblender}, we handle the foreground and background separately to prevent the pixel mixing problem around the object boundary. 





\section{Experiments}\label{sec:exp}
We evaluate our method with various dynamic scenes.

\begin{wrapfigure}{l}{0.182\textwidth}
\vspace{-8mm}
  \begin{center}
    \includegraphics[width=0.21\textwidth]{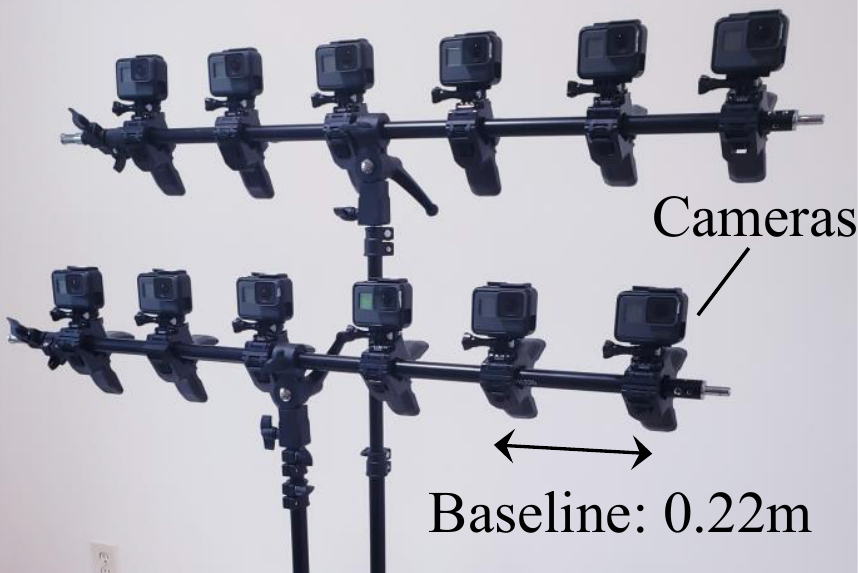}
  \end{center}
    \vspace{-6mm}
  \caption{\small Camera rig.} 
  \label{Fig:rig}
  \vspace{-2mm}
\end{wrapfigure}

\begin{table*}[t]
\centering
\small
\begin{tabular}{|l|c|c|c|c|c|c|c|c|c|c}
\hline

 F+B / F-only \ \ $\searrow$& \scriptsize{Jumping}  & \scriptsize{Skating}  & \scriptsize{Truck}  & \scriptsize{DynaFace} &  \scriptsize{Umbrella} & \scriptsize{Balloon1} & \scriptsize{Balloon2}  & \scriptsize{Teadybear}  & \scriptsize{Avg.}   \\
\hline
\scriptsize{MVS}~\cite{schonberger2016pixelwise} &\scriptsize{0.53 / 2.12} &\scriptsize{0.29 / 6.81}  &\scriptsize{0.52 / 2.94}   &\scriptsize{0.05 / 0.21}    &  \scriptsize{\textbf{0.35} / 4.70}&\scriptsize{0.13 / 1.72}  &  \scriptsize{\textbf{0.04} / 0.31}&  \scriptsize{\textbf{0.06} / 0.92}&\scriptsize{0.24 / 2.46}     \\
\hline

\scriptsize{RMVSNet}~\cite{yao2019recurrent}&\scriptsize{0.61 / 1.55}  &\scriptsize{0.76 / 1.56}   &\scriptsize{0.84 / 2.43}  &\scriptsize{2.24 / 1.57}  &\scriptsize{0.67 / 5.24}  &\scriptsize{0.23 / 1.40}  &\scriptsize{0.13 / 0.38}  &\scriptsize{0.58 / 0.89}  & \scriptsize{0.75 / 1.87} \\

\hline
\scriptsize{MonoDepth}~\cite{lasinger2019towards}&\scriptsize{1.79 / 2.55}  &\scriptsize{1.34 / 2.02} &\scriptsize{2.62 / 3.86}  &\scriptsize{0.39 / 0.74}  &  \scriptsize{2.69 / 4.75}  &\scriptsize{1.07 / 1.88}  &\scriptsize{1.06 / 0.99}  &\scriptsize{0.76 / 0.28}   &\scriptsize{1.46 / 2.13} \\ 
\hline

\scriptsize{Sparse2Dense}~\cite{mal2018sparse}&\scriptsize{1.35 / 3.26}  &\scriptsize{1.35 / 10.66}  & \scriptsize{2.15 / 7.60} & \scriptsize{0.20 / 0.34} &   \scriptsize{1.35 / 6.40} & \scriptsize{0.53 / 3.03} & \scriptsize{0.48 / 0.65} & \scriptsize{0.32 / 0.90} & \scriptsize{0.96 / 4.10} \\ 
\hline
\hline

\scriptsize{DFNet-$L_{g}$}& \scriptsize{1.26 / 1.31} &\scriptsize{0.81 / 0.76}  & \scriptsize{1.60 / {1.24}} & \scriptsize{0.26 / 0.91} &   \scriptsize{2.19 / {1.98}} & \scriptsize{0.93 / 1.36} & \scriptsize{0.53 / 0.30} & \scriptsize{1.91 / 0.97} &\scriptsize{1.18 / 1.10} \\ 
\hline
\scriptsize{DFNet-$L_{l}$}& \scriptsize{0.46 / 1.58} & \scriptsize{0.15 / 1.38}& \scriptsize{0.62 / 3.34} & \scriptsize{0.09 / 0.26} &   \scriptsize{0.58 / 3.14} & \scriptsize{0.15 / 1.57} & \scriptsize{0.08 / 0.30} & \scriptsize{0.16 / 0.67} &\scriptsize{0.28 / 1.53}  \\ 
\hline
\scriptsize{DFNet-$L_{e}$}& \scriptsize{0.38 / 0.93} & \scriptsize{0.14 / 0.47}&  \scriptsize{0.52 / 1.09}&  \scriptsize{0.07 / 0.12}&    \scriptsize{0.52 / 2.48}&  \scriptsize{0.15 / 1.20}& \scriptsize{0.06 / 0.24} & \scriptsize{0.17 / 0.48} & \scriptsize{0.26 / 0.87} \\ 
\hline
\scriptsize{DFNet-$L_{s}$}& \scriptsize{0.37 / 1.09}  &\scriptsize{0.14 / 0.51} &\scriptsize{0.53 / 1.11}  &\scriptsize{0.07 / 0.13}   &\scriptsize{0.59 / 2.54}  &\scriptsize{0.16 / 1.18}  &\scriptsize{0.07 / 0.25}  &\scriptsize{0.16 / 0.52}  &\scriptsize{0.26 / 0.91} \\ 
\hline
\scriptsize{DFNet}&  \scriptsize{\textbf{0.35} / \textbf{0.76}} & \scriptsize{\textbf{0.12} / \textbf{0.40}}& \scriptsize{\textbf{0.41} / \textbf{0.83}} & \scriptsize{\textbf{0.03} / \textbf{0.08}}  &\scriptsize{0.37 / \textbf{1.90}}  &\scriptsize{\textbf{0.12} / \textbf{1.11}}  &\scriptsize{0.05 / \textbf{0.23}}  &\scriptsize{0.17 / \textbf{0.32}}  &\scriptsize{\textbf{0.20} / \textbf{0.70}}\\ 
\hline
\end{tabular}
\vskip-5pt
\caption{\small Results of quantitative evaluation for the task of depth estimation from dynamic scenes. RMSE in the metric scale is used for evaluation. F and B represent the foreground and background, respectively. The lower is the better.}
\label{table:depth_tab}
\vskip-3pt
\end{table*}

\begin{figure*}[t]
	\begin{center}
	\includegraphics[width=1\textwidth]{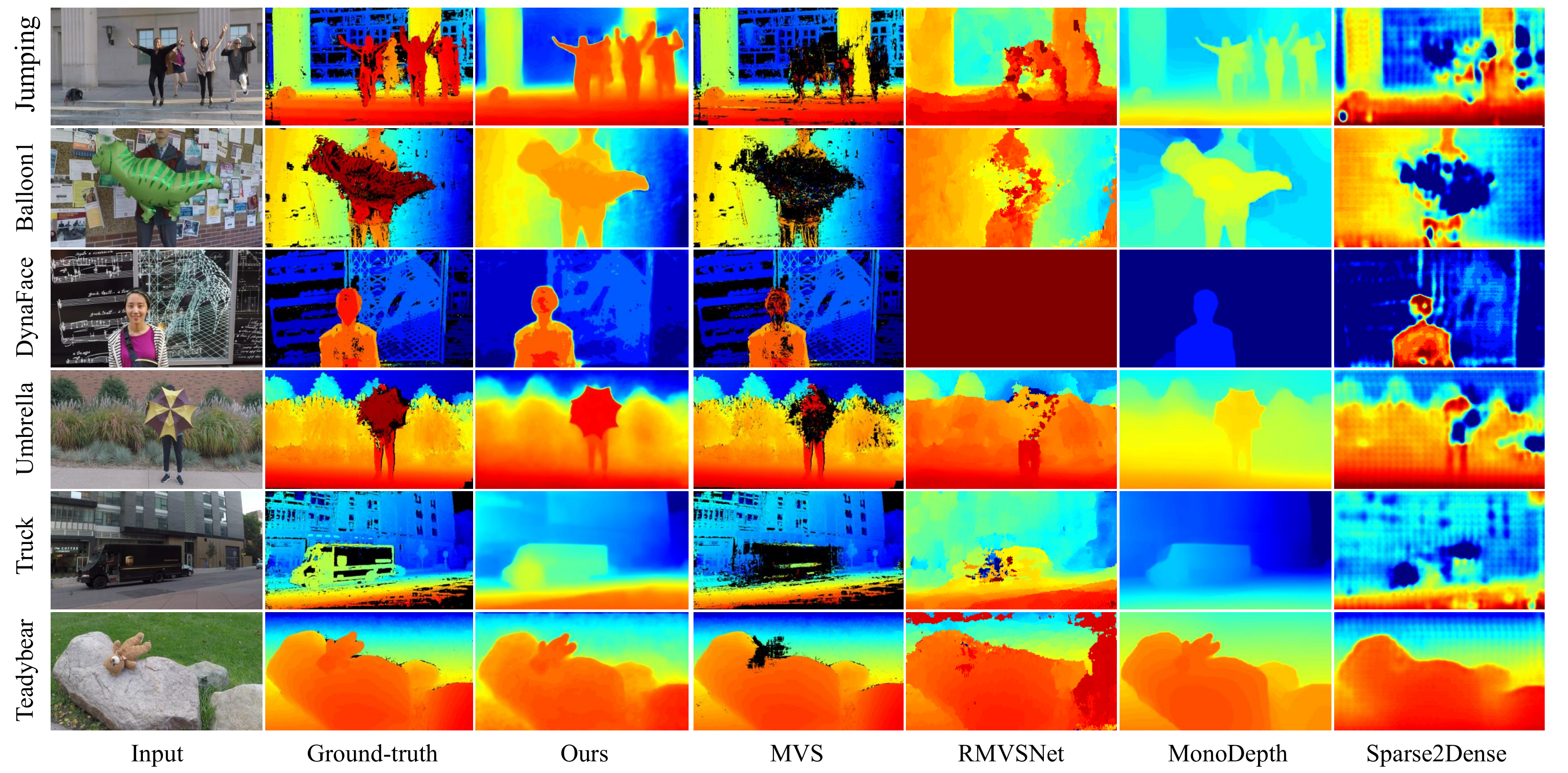}
	\end{center}	
	\vspace{-5mm}
	\caption{\small Qualitative comparison of the dynamic scene depth estimation from each method.}
	\label{fig:depth_res}
\end{figure*}
\noindent\textbf{Dynamic Scene Dataset} We collect dynamic scenes using two methods. (1) Moving monocular camera: short-term dynamic events ($\sim$ 5s) are captured by a hand-held monocular moving camera (Samsung Galaxy Note 10) with 60Hz framerate and 1920$\times$1080 resolution. We sub-sample the sequence if the object motion is not salient, and therefore, the degree of the scene motion is significantly larger than that of the camera egomotion where quasi-static dynamic reconstruction does not apply. Four dynamic scenes are captured, which includes human activity, human-object interaction, and animal movement (see the supplementary video). These scenes are used for the qualitative evaluation, where we use half-resolution inputs. (2) Stationary multiview cameras: 8 scenes are captured by a static camera rig with 12 cameras (GoPro Black Edition), where the ground truth of depth estimation and view synthesis are available for the quantitative evaluation. The cameras are located at two levels, and at each level, 6 cameras are evenly distributed with 0.22m baseline as shown in Figure~\ref{Fig:rig}. All cameras are manually synchronized. The dataset is categorized into following: (1) Human: a single or multiple people show their dynamic motion, e.g., dynamic facial expression and body motion. (2) Interaction: a person interacts with objects, e.g., umbrella, balloon, and skate. (3) Vehicle: a truck rigidly move from the right side of the road to the left. (4) Stop motion: a doll is sequentially captured in the different location. When testing, we use a set of images sampled from each camera at different time instant to simulate a moving monocular camera. Given the set of collected images, we calibrate the intrinsic and extrinsic parameters of the moving camera using structure-from-motion~\cite{schoenberger2016sfm}. 

\noindent\textbf{Quantitative Evaluation Metric} We evaluate the accuracy of depth estimation and view synthesis using the multiview dataset. (1) Depth estimation: given the estimated depth, we measure root mean square error (RMSE) by comparing to the ground-truth depth computed by multiview stereo. The error is represented in metric scale (m), i.e., the scale of the estimated depth is upgraded to the metric space using the physical length of the camera baseline. We exclude the region that cannot be reconstructed by multiview stereo. 
(2) View synthesis: we measure the mean of the optical flow magnitude from the ground-truth image to the synthesized one to validate the view invariant property of the depth map. Ideally, it should be close to 0 with the perfect depth map. Additionally, we measure the perceptual similarity~\cite{zhang2018unreasonable} (i.e., the distance of VGG features) with the ground-truth to evaluate the visual plausibility of the synthesized view, where its range is normalized into [0, 1] (the lower is the better).


\noindent\textbf{Baselines and Ablation Study} We compare our depth estimation and view synthesis methods with a set of baseline approaches. 
For the depth evaluation, we compare our method with four baselines: 
1) Multiview stereo (MVS~\cite{schonberger2016pixelwise}) assumes that a scene is stationary. For the pixel of which MVS failed to measure the depth, we assign the average of valid depth. 2) RMVSNet~\cite{yao2019recurrent} is a learning based multiview stereo algorithm. 3) MonoDepth~\cite{lasinger2019towards} predicts the depth from a single view image. As it produces the normalized depth, we re-scale the predicted depth by using the mean and standard deviation from MVS depth. 4) Sparse2Dense~\cite{mal2018sparse} completes the depth given an incomplete depth estimation, where we use MVS depth as an input. As this method requires the metric depth, we upgrade the estimated depth to the metric space using the physical length of the camera baseline. In conjunction with comparative evaluations, we conduct an ablation study to validate the choice of losses.


For the view synthesis evaluation, we compare our view warping method (bi-directional 3D warping) with as-similar-as-possible warping~\cite{liu2013bundled} which warps an image by estimating grid-wise affine transforms. The correspondences of the warping are computed by projecting the estimated depth, i.e., transporting pixels in a source image to a novel view through the view-invariant depth. In Table~\ref{table:viewsynthesis_experiment}, we denote bi-directional warping followed by the DeepBlender refinement as B3W, and as-similar-as-possible warping followed by the DeepBlender as ASAPW. Note that the DeepBlender refinement is applied to all methods except for DFNet+B3W-DeepBlender which evaluates the effect of the refinement by eliminating the DeepBlender. On top of the comparison with different warping methods, we also test all possible combination of depth estimation methods with view warping methods as listed in Table~\ref{table:viewsynthesis_experiment}. It quantifies how the quality of depth maps affect the view synthesis results.





\noindent\textbf{Dynamic Scene Depth Estimation}
In Table~\ref{table:depth_tab}, we summarize the accuracy of dynamic scene depth estimation results evaluated on: 1) the entire scene, and 2) the only dynamic contents. For the entire scene, our method shows the best results on average, followed by MVS with 0.04 m accuracy gap. In the sequence of umbrella and teadybear, MVS shows the better accuracy for the entire scene than ours due to the highly occupant background area as shown in Figure~\ref{fig:depth_res}, i.e., the depth estimation of dynamic contents much less contributes to depth accuracy evaluation than one of the background. From the evaluation on the only dynamic contents, our method (DFNet) also shows the best result with the noticeable accuracy improvement (1.17 m) from the second best method (MonoDepth). 


While the relative depth of MonoDepth is well reflective of the ground-truth, its depth range is often biased to a specific range, e.g., the foreground object is located much closer to the background scenes. Sparse2Dense does not fully reconstruct the background depth even with the MVS depth as inputs, and the predicted foreground depth is completely incorrect. It indicates that fusing the individual strength  of learning-base and stereo-based geometry is essential to obtain the globally coherent and complete depth map from dynamic scenes. From Figure~\ref{fig:depth_res}, we can further notice that the learning based multiview stereo (RMVSNet) also fail to model the dynamic foreground geometry. In our experiment, RMVSNet completely fail when the object is too close to the camera.  

From the ablation study described in Table~\ref{table:depth_tab}, $L_g$ is the most critical self-supervision signal as the MVS depth plays the key role to convey the accurate static depth. Those accurate depths play the fiducial point for the other self-supervision signals to predict the depth on the missing area. From DFNet-$L_l$, we can verify that the single view depth estimation can upgrade the depth accuracy around the dynamic contents by guiding it with accurate relative depths. Although the contribution of $L_e$ and $L_s$ are relatively small than others, it helps to regularize the object scene motion and the spatial smoothness of the foreground depth which are keys to reduce the artifacts of the novel view synthesis. 

\begin{figure}[t]
\vspace{2mm}
	\begin{center}
\includegraphics[width=1\linewidth]{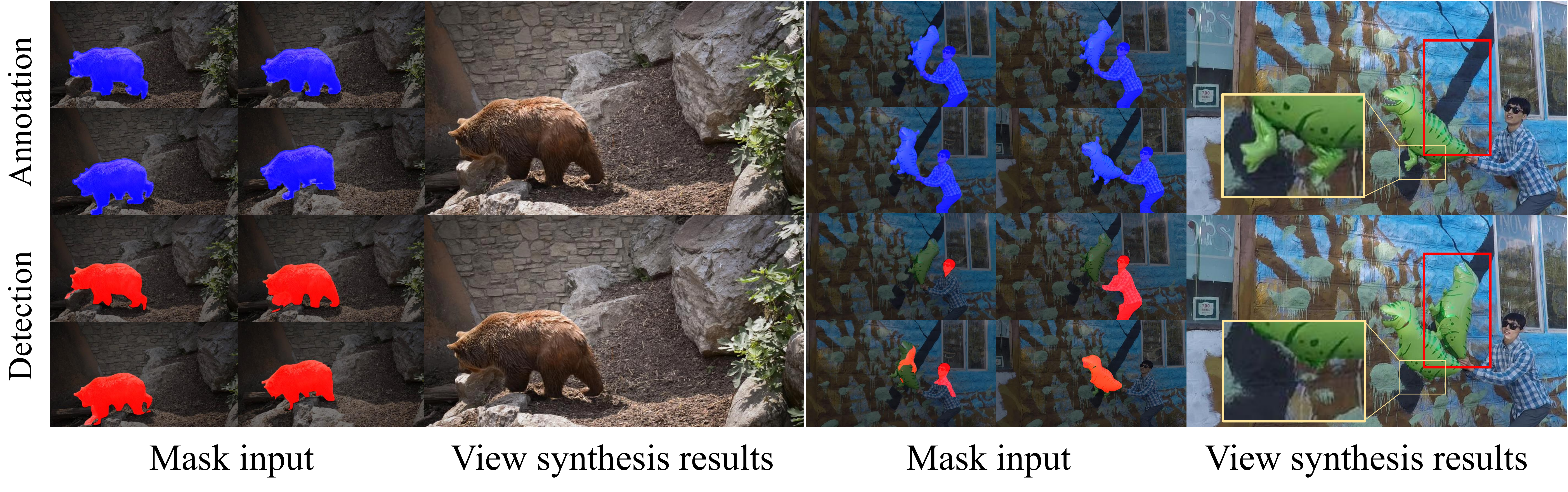}
	\end{center}
\vspace{-5mm}
    \caption{ \small{The mask detection with small mistakes (left) does not have a significant impact on the view synthesis results. However, if the mask detection is completely failed (right), it produces artifacts such as object fragmentation (yellow box) or afterimage (red box).}}
    \label{fig:mask}
\end{figure}
\begin{figure*}[t]
	\begin{center}
		\includegraphics[trim={0 12cm 0 0},clip,width=6.7in]{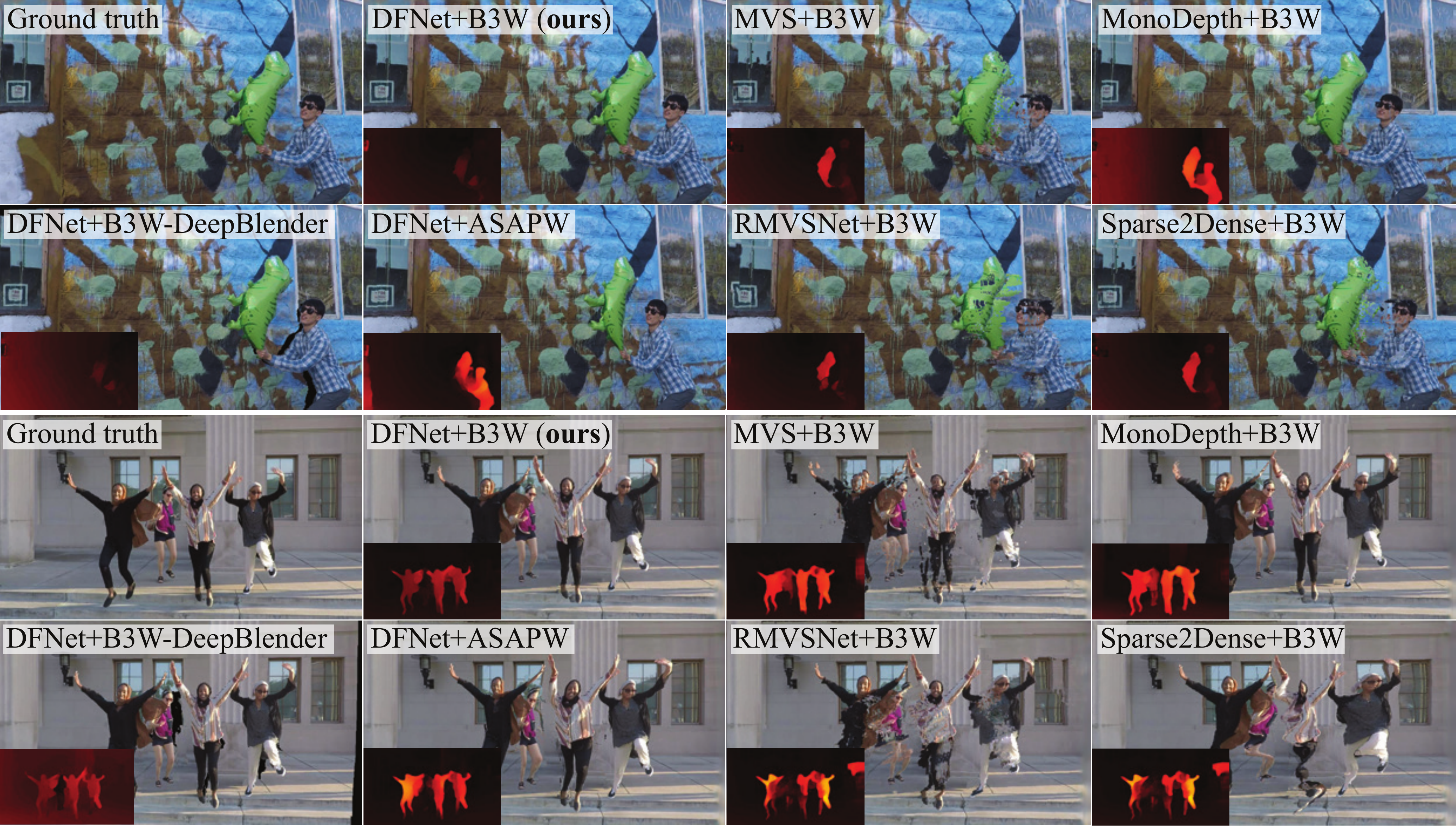}
	\end{center}	
	\vspace{-6mm}
	\caption{\small Qualitative comparison on the view synthesis task.  The pixel error is shown in the inset image (maximum pixel error is set to 50 RGB distance). }
	\label{fig:qual_viewsynth}
\vspace{-2mm}
\end{figure*}

\noindent\textbf{Dynamic Scene Novel View Synthesis}
Table~\ref{table:viewsynthesis_experiment} shows the quantitative evaluation of view synthesis, and the associated qualitative results are shown in Figure~\ref{fig:qual_viewsynth}. From the qualitative results, we can notice that two types of artifacts can be produced depending on the warping methods: B3W produces flying pixel noises, i.e., a pixel is floating due to the warping with incorrect depths, while ASAPW produces image distortion. Such artifacts lead to the increase of the perceptual distance with ground-truths as it captures the structural similarity. On average, our method (DFNet+B3W) shows the smallest perceptual distance (0.15), indicating that the geometry from our depth map is highly preservative of scene structure. The comparison of DFNet+B3W with DFNet+ASAPW demonstrates that, given an accurate depth map, pixel-wise warping (B3W) is the better choice over the grid-wise warping (ASAWP) for view synthesis. From the results of DFNet+B3W-Deepblender, we can observe the large improvement of perceptual similarity compared to the results without DeepBlender, indicating that the refinement step (hole filling and noise reduction) is essential for visual plausibility.

Our method (DFNet+B3W) performs the best even for the flow evaluation (5.3 pixels). MVS+B3W is following our method with the 6.8 pixel errors but it produces a significant pixel noise around the dynamic contents as shown in Figure~\ref{fig:qual_viewsynth}. While MonoDepth+B3W reconstructs visually plausible results in Figure~\ref{fig:qual_viewsynth}, it accompany with large flow errors (10.8 pixels on average), meaning that this result is not geometrically plausible. Note that, the optical flow error of DFNet+B3W-Deepblender is much higher than DFNet+B3W because the flow estimation algorithm~\cite{Sun2018PWC-Net} shows significant confusion when there are holes around the image boundary and dynamic contents.

\begin{table*}[t]
\centering
\scriptsize
\begin{tabular}{|l|c|c|c|c|c|c|c|c|c|}
\hline

Perceptual Sim. / Optical Flow  $\searrow$ & \scriptsize{Jumping} & \scriptsize{Skating}  & \scriptsize{Truck} & \scriptsize{DynaFace}  & \scriptsize{Umbrella} & \scriptsize{Balloon1} & \scriptsize{Balloon2}  & \scriptsize{Teadybear}  & \scriptsize{Avg.} \\
\hline
\scriptsize{MVS}~\cite{schonberger2016pixelwise}+ASAPW~\cite{liu2013bundled} &0.21 / 7.0 & 0.17 / 9.3 & 0.10 / 4.0 & 0.30 / 19.0 & 0.19 / 7.5&0.23 / 16.0  & 0.17 / 6.7  &1.80 / 4.9 &0.19 / 9.3 \\
\hline
\scriptsize{RMVSNet}~\cite{yao2019recurrent}+ASAPW~\cite{liu2013bundled}& 
0.22 / 6.4 & 0.23 / 13.1 & 0.11 / 3.4 & 0.98 / 10.2 & 0.19 / 7.2 &0.23 / 14.9  &0.16 / 6.3  & 0.20 / 10.0 &0.29 / 8.9  \\ 
\hline
\scriptsize{MonoDepth}~\cite{lasinger2019towards}+ASAPW~\cite{liu2013bundled}&0.23 / 9.1 & 0.18 / 11.8 & 0.10 / 5.1 & 0.32 / 20.9 &0.20 / 9.8  &0.25 / 17.3 &0.23 / 11.4 &0.17 / 7.8 &0.20 / 11.7  \\ 
\hline
\scriptsize{Sparse2Dense}~\cite{mal2018sparse}+ASAPW~\cite{liu2013bundled}&0.23 / 7.5 & 0.19 / 9.4 & 0.11 / 4.8  & 0.31 / 20.8 & 0.19 / 7.0 &0.23 / 13.7 &0.16 / 6.6 &0.19 / 6.4 &0.20 / 9.52 \\ 
\hline

\hline
\scriptsize{MVS}~\cite{schonberger2016pixelwise}+B3W & 0.24 / 7.0 & 0.20 / 9.2 & 0.12 / 3.5 & 0.27 / 7.5 &  0.19 / 5.7 & 0.23 / 14.4 & 0.17 / 5.4 &   0.13 / \textbf{1.5}   &0.19 / 6.8 \\
\hline
\scriptsize{RMVSNet}~\cite{yao2019recurrent}+B3W  & 0.23 / 5.6 & 0.23 / 14.8 & 0.14 / 3.3 & 1.0 / 10.8   & 0.19 / 5.6 & 0.23 / 12.0 & 0.16 / 5.1  & 0.19 / 8.9 & 0.29 / 8.2\\ 
\hline
\scriptsize{MonoDepth}~\cite{lasinger2019towards}+B3W & 0.23 / 8.5  & 0.18 / 11.4 & 0.10 / 5.0 & 0.32 / 19.1 &  0.19 / 8.5 & 0.24 / 17.3 & 0.23 / 11.4 &  0.15 / 5.2 & 0.20 / 10.8\\ 
\hline
\scriptsize{Sparse2Dense}~\cite{mal2018sparse}+B3W& 0.24 / 7.3  & 0.20 / 9.2 & 0.13 / 4.7 & 0.31 / 11.7 &   0.2 / 6.7 & 0.24 / 14.0 &  0.18 / 6.6 & 0.17 / 4.8 &0.22 / 8.12 \\ 
\hline

\hline
\scriptsize{DFNet+ASAPW}~\cite{liu2013bundled}&
0.20 / 5.8 & 0.17 / 9.3 & 0.09 / 3.0 & 0.30 / 18.0 & 0.18 / 6.4 & 0.20 / 13.3 &0.16 / 6.4 & 0.17 / 5.8 & 0.18 / 8.5  \\ 
\hline
\scriptsize{DFNet+B3W-DeepBlender}&  0.23 / 8.2&  0.21 / 13.1 &   0.12 / 4.8& 0.30 / 15.6 &   0.22 / 9.0 & 0.25 / 15.8 &  0.20 / 9.2 & 0.18 / 4.7 & 0.21 / 10.1 \\ 
\hline
\scriptsize{DFNet+B3W (\textbf{ours})}& \textbf{0.16} / \textbf{4.2} & \textbf{0.15} / \textbf{8.8} & \textbf{0.08} / \textbf{2.5} & \textbf{0.22} / \textbf{6.2} &  \textbf{0.16} / \textbf{3.6} &  \textbf{0.18} / \textbf{10.6}  & \textbf{0.14} / \textbf{5.1} &  \textbf{0.13} / 2.0  &\textbf{0.15} / \textbf{5.3}\\ 
\hline
\end{tabular}
\vspace{-2mm}
\caption{\small Quantitative evaluation results on the dynamic scene novel view synthesis task. To measure the accuracy, we compute perceptual similarity and optical flow magnitude between the ground-truth and the synthesized image. }
\vspace{0mm}
\label{table:viewsynthesis_experiment}
\end{table*}

\noindent\textbf{Limitation}
It is worth noting a few limitations of our method. The DFNet may not perform well when a viewing angle between neighboring views are larger (e.g., rotating more than $45^{\circ}$), which may decrease the amount of overlaps of dynamic contents. If the scene is highly cluttered by many objects from both background and foreground (e.g., many people, thin poles, and trees), our pipeline could cause noisy warping results due to the significant depth discontinuities from the clutter. Our method will fail in the scenes where the camera calibration does not work, e.g., a scene largely occupied by dynamic contents~\cite{lv2018learning}. Finally, our view synthesis with completely failed foreground mask produces significant artifacts such as afterimages and object fragmentation as shown in Figure~\ref{fig:mask}.

\section{Conclusion}
\label{sec:conclusion}
In this paper, we study the problem of monocular view synthesis of a dynamic scene. The main challenge is to reconstruct dynamic contents to produce geometrically coherent view synthesis, which is an ill-posed problem in general. To address this challenge, we propose to learn a scale correction function that can upgrade the depth from single view (DSV), which allows matching to the depth of the multi-view solution (DMV) for static contents while producing locally consistent scene motion. Given the computed depth, we synthesize a novel view image using the DeepBlender network that is designed to combine foreground, background, and missing regions. Through the evaluations for depth estimation and novel view synthesis, we demonstrate that the proposed method can apply to the daily scenario captured from a monocular camera. \\

\noindent \textbf{Acknowledgement} This work was partly supported by the NSF under IIS 1846031 and CNS 1919965.


{\small
\bibliographystyle{ieee_fullname}
\bibliography{egbib}
}

\end{document}